\documentclass[a4paper,conference]{IEEEtran}
\usepackage[left=1.57cm,right=1.57cm,top=0.95cm,bottom=2.54cm]{geometry}

\usepackage[american]{babel}
\usepackage{hyphenat}
\usepackage{booktabs}

\usepackage{algorithmic}
\usepackage{algorithm}
\usepackage{array}
\usepackage{soul}

\usepackage{multirow}

\usepackage{subfigure}

\usepackage{textcomp}
\usepackage{url}
\usepackage{verbatim}
\usepackage{graphicx}
\usepackage{xcolor}

\usepackage{stfloats}

\usepackage{cite}
\usepackage[utf8]{inputenc}
\usepackage[T1]{fontenc}
\usepackage{textcomp}
\usepackage{microtype}
\usepackage{calc}
\usepackage[normalem]{ulem}

\usepackage{amsmath}
\usepackage{amssymb}
\usepackage{amsfonts}
\usepackage{amsthm}

\usepackage{mathtools}

\usepackage[capitalize,noabbrev,nameinlink]{cleveref}

\RequirePackage{xstring}
\RequirePackage{xparse}
\RequirePackage[]{acro}
\makeatletter
\@ifpackagelater{acro}{2019/02/17}{
	\NewDocumentCommand\acrodef{mO{#1}mG{}}{\DeclareAcronym{#1}{short={#2}, long={#3}, foreign-plural={}, #4}}
}{
	\NewDocumentCommand\acrodef{mO{#1}mG{}}{\DeclareAcronym{#1}{short={#2}, long={#3}, #4}}
}
\makeatother
\acrodef{AIFS}{Arbitration Inter-Frame Spacing}
\acrodef{AoI}{Age of Information}

\def\todoCtd#1{%
	TODO: #1%
	\ifx&#1&.\fi%
	\endgroup
	\cbend
	\relax
}

\begin{document}

\title{Towards Multi-Model LLM Schedulers: Empirical Insights into Offloading and Preemption}

\author{%
  \IEEEauthorblockN{%
    Mert Yildiz\IEEEauthorrefmark{1},
    Pietro Spadaccino\IEEEauthorrefmark{1},
    Alexey Rolich\IEEEauthorrefmark{1},
    Francesca Cuomo\IEEEauthorrefmark{1},
    Andrea Baiocchi\IEEEauthorrefmark{1}
  }
  \IEEEauthorblockA{%
    \IEEEauthorrefmark{1}Sapienza University of Rome, Rome, Italy\\
    Email: \{mert.yildiz, pietro.spadaccino, alexey.rolich, francesca.cuomo, andrea.baiocchi\}@uniroma1.it
  }
}

\maketitle

\begin{figure*}[b]
\centering
\noindent\fbox{%
\begin{minipage}{\dimexpr\linewidth-2\fboxsep-2\fboxrule\relax}
\footnotesize
\noindent \copyright\ 2026 IEEE. Personal use of this material is permitted. Permission from IEEE must be obtained for all other uses, in any current or future media, including reprinting/republishing this material for advertising or promotional purposes, creating new collective works, for resale or redistribution to servers or lists, or reuse of any copyrighted component of this work in other works.

\vspace{0.15cm}
\hrule height 0.4pt 
\vspace{0.15cm}

\noindent \textit{2026 Mediterranean Artificial Intelligence and Networking Conference (MAIN 2026)}
\end{minipage}}
\end{figure*}

\begin{abstract}
Modern deployments of Large Language Models (LLMs) increasingly require serving multiple models with diverse architectures, sizes, and specialization on shared, heterogeneous hardware. This setting introduces new challenges for resource allocation, dispatching, and scheduling, particularly under GPU memory constraints where partial CPU-GPU offloading and preemption become necessary. While existing systems primarily optimize throughput for a single model, comparatively little work addresses multi-model scheduling under these conditions. In this paper, we present an empirical study of how different LLMs behave across hardware platforms, focusing on the performance implications of layer offloading and preemption. We show that offloading leads to strongly non-linear and model-dependent degradation in decode throughput, with smaller models exhibiting sharper sensitivity to reduced GPU residency. We further demonstrate that preemption incurs substantial overhead, largely dominated by model state reload rather than key-value cache transfer, and that this cost varies significantly across models and hardware platforms. Additionally, we highlight the role of sequence length and interconnect bandwidth in amplifying data movement and execution inefficiencies. Based on these findings, we identify a set of key features that future schedulers must consider, including model-specific offloading sensitivity, workload characteristics, and the cost structure of preemption and data transfer. These insights provide guidance for the design of next-generation LLM serving systems capable of efficiently managing heterogeneous, multi-model workloads with hybrid CPU-GPU execution.
\end{abstract}


\acresetall
\IEEEpeerreviewmaketitle

%

\section{Introduction}
\label{sec:intro}

In recent years, large language models (LLMs) have become a central component of modern artificial intelligence systems, enabling advanced capabilities in natural language understanding, generation, and reasoning. The rapid pace of model development has led to a diverse ecosystem of LLMs that differ in architecture, size, training data, and performance characteristics. As a result, real-world deployments increasingly involve not a single model, but a collection of heterogeneous models that are continuously updated, replaced, or specialized for different tasks.

In this field, practitioners routinely evaluate and deploy multiple models to balance trade-offs between accuracy, latency, and resource consumption. In both research and industrial settings, it is common to select different models depending on application requirements, system constraints, or cost considerations. Moreover, the ecosystem of LLMs is increasingly specialized: some models are optimized for natural language generation, others for code synthesis, and others for agentic or tool-augmented reasoning. This diversity motivates the need for systems capable of serving multiple models concurrently, as no single model uniformly dominates across all tasks and deployment scenarios.

Managing such heterogeneous workloads requires efficient mechanisms for resource allocation, request dispatching, and scheduling across available hardware. These challenges are exacerbated in environments with limited GPU resources, where multiple models compete for memory and compute, and where dynamic workload patterns introduce variability in demand. Effective system design must therefore account for both inter-model heterogeneity and intra-model execution characteristics, while ensuring high utilization and predictable latency.

Existing LLM serving systems have primarily focused on optimizing throughput for a single model by batching and executing multiple requests in parallel. Systems such as vLLM \cite{kwon2023vllm} and related frameworks exploit techniques like continuous batching and efficient key-value cache management to maximize GPU utilization when serving many concurrent requests to the same model. While these approaches are effective in homogeneous settings, they do not directly address scenarios in which requests are distributed across multiple models with different resource footprints and execution profiles.

In contrast, comparatively little attention has been devoted to the problem of serving multiple LLMs concurrently within a shared infrastructure. In such settings, scheduling decisions must consider not only the number of requests, but also model-specific characteristics, partial offloading strategies, and the interaction between CPU and GPU resources. This introduces a fundamentally different set of trade-offs, where isolating models may simplify execution but reduce utilization, while sharing resources across models increases efficiency at the cost of greater scheduling complexity.

In this paper, we investigate the behavior of diverse LLMs, spanning different architectures and model scales, across heterogeneous hardware platforms. Our goal is to characterize how these models interact with system resources under varying execution configurations, including partial CPU-GPU offloading. Through this analysis, we identify the key factors that influence performance, memory usage, and latency across models and hardware settings. Building on these insights, we outline the requirements for next-generation scheduling systems, highlighting the considerations necessary to support advanced mechanisms such as preemption and hybrid CPU-GPU offloading. Our results provide guidance for the design of future schedulers capable of efficiently managing multi-model workloads in dynamic and resource-constrained environments.

\subsection{LLM Computing Model}

Large Language Models based on transformer architectures execute inference as a sequence of layered computations, where each layer performs a set of operations including linear projections, attention mechanisms, and non-linear transformations. These layers are strictly ordered and must be evaluated sequentially due to data dependencies between them. In heterogeneous environments, individual layers can be mapped either to GPUs or CPUs. GPUs provide significantly higher throughput due to massive parallelism and specialized tensor cores, whereas CPUs offer lower computational performance but greater flexibility and memory capacity. Consequently, partial offloading strategies assign subsets of layers to the CPU to alleviate GPU memory pressure, at the cost of increased latency due to slower execution and potential data transfer overheads.

Inference in LLMs follows an autoregressive generation process, in which tokens are produced sequentially. At each decoding step, the model consumes the previously generated tokens and computes the probability distribution for the next token. This introduces an inherent temporal dependency across steps, preventing parallelization across tokens during generation. As a result, end-to-end latency is determined not only by the per-layer execution time but also by the cumulative effect of sequential token generation, making scheduling decisions sensitive to both computation placement and token-level iteration dynamics.

Memory consumption in LLM inference can be decomposed into a static and a dynamic component. The static component corresponds to model parameters, which remain fixed during inference and must reside in memory across all decoding steps. The dynamic component arises from intermediate activations and, critically, the key-value (KV) cache used in attention mechanisms. The KV cache stores representations of previously processed tokens to avoid redundant computation in subsequent steps, leading to memory growth proportional to the sequence length. This dynamic memory footprint significantly impacts resource allocation and scheduling, particularly when multiple concurrent inference requests are served or when operating under constrained GPU memory, thereby motivating hybrid CPU-GPU execution strategies.





Our contributions are as follows:
\begin{enumerate}
    \item we conduct a systematic empirical evaluation of state-of-the-art LLMs across heterogeneous hardware platforms, demonstrating that key performance trends, including the impact of offloading and scaling behavior, remain consistent across models and systems;
    \item we analyze the performance implications of partial CPU-GPU offloading and preemption, quantifying their effects on decode throughput, latency, and data movement, and identifying the dominant sources of overhead;
    \item we distill these observations into a set of key features that future schedulers must consider, providing guidance for the design of scheduling policies that support multi-model workloads with hybrid CPU-GPU execution and preemption.
\end{enumerate}

The remainder of this paper is organized as follows. \Cref{sec:relwork} reviews related work. \Cref{sec:setup} describes the experimental setup. \Cref{sec:offloading,sec:migrating} present the offloading and preemption results respectively. \Cref{sec:discussion} distills the findings into key features for future schedulers, and conclusions are drawn in \Cref{sec:conclusions}.

%
\section{Related Work}
\label{sec:relwork}

Several systems address the case where model weights do not fit in GPU memory.
FlexGen~\cite{sheng2023flexgen} formulates tensor placement across GPU, CPU, and disk as a linear program that maximizes batch throughput.
LIA~\cite{kim2025lia} leverages Intel AMX matrix-multiplication units on the CPU to cooperatively execute sublayers alongside the GPU, combined with CXL memory offloading, achieving up to 19$\times$ lower latency than prior offloading frameworks on OPT-175B.
PowerInfer~\cite{song2024powerinfer} exploits activation sparsity to keep only frequently activated neurons on the GPU.
Fiddler~\cite{kamahori2024fiddler} and KTransformers~\cite{chen2025ktransformers} extend heterogeneous execution to Mixture-of-Experts architectures.
NEO~\cite{neo2025} offloads attention and KV-cache operations to the CPU to relieve memory pressure during online serving, and LM-Offload~\cite{wu2024lmoffload} builds an analytical model to guide placement decisions.
Each of these works evaluates its design at a handful of fixed configurations, typically comparing against a fully-GPU and a fully-CPU baseline.
None systematically varies the fraction of offloaded layers as an independent variable, leaving open the question of how decode throughput degrades as a continuous function of the offloading ratio across different model sizes and hardware platforms.

Managing the KV cache, whose memory footprint grows linearly with sequence length~\cite{hooper2024kvquant}, and deciding when to preempt or reschedule requests are tightly coupled problems that have driven much of the recent progress in LLM serving.
Orca~\cite{yu2022orca} introduces continuous batching, allowing requests to enter and leave a running batch at every decoding step.
vLLM~\cite{kwon2023vllm} manages KV caches via paged memory inspired by operating-system virtual memory, and when memory runs out, preempts requests by discarding their cache and later recomputing it.
Sarathi~\cite{agrawal2023sarathi,agrawal2024sarathiserve} piggybacks decode tokens onto prefill batches through chunked prefill, improving GPU utilization by mixing the two phases within a single iteration.
Splitwise~\cite{patel2024splitwise} and DistServe~\cite{zhong2024distserve} disaggregate prefill and decode onto separate GPU pools to match each phase to its bottleneck resource, which requires migrating the KV cache over the network after prompt processing completes.
D\'{e}j\`{a}Vu~\cite{strati2024dejavu} streams KV-cache state between GPU and CPU or across nodes to enable fast recovery from failures and low-overhead migration in distributed serving.
InfiniGen~\cite{lee2024infinigen} offloads KV-cache entries to CPU memory and speculatively prefetches only the entries needed for the next decoding step back to the GPU.
InstInfer~\cite{pan2024instinfer} pushes this further by offloading attention computation itself to CXL-attached storage devices, keeping the GPU free for other operations.
Llumnix~\cite{sun2024llumnix} introduces live migration of KV caches across GPU instances for runtime rescheduling and load balancing, analogous to virtual-machine live migration.
FastSwitch~\cite{shen2024fastswitch} identifies that preemption-induced context switching can stall inference for longer than the useful computation itself, and proposes coarser-grained batched transfers to reduce this overhead.

FastServe~\cite{wu2023fastserve} swaps KV-cache state to host memory over PCIe and schedules requests with a skip-join Multi-Level Feedback Queue that supports token-level preemption.
S$^3$~\cite{jin2023s3} and Zheng et al.~\cite{zheng2024response} predict output lengths to approximate shortest-job-first ordering without preemption.
Kim et al.~\cite{kim2024dbms} adapt DBMS-inspired cost models to LLM inference and find that strategically preempting short requests under memory pressure can up to double throughput compared to preemption-avoidance policies.
SpotServe~\cite{miao2024spotserve} addresses preemption at a coarser grain, migrating entire inference jobs across preemptible cloud instances.

All of the systems discussed above assume that a single model occupies the GPU throughout the serving lifetime.
In practice, inference clusters must serve requests targeting many different models, and multiplexing a GPU across models remains an open challenge that recent surveys identify as central to scalable LLM deployment~\cite{heisler2025llmschedsurvey}.
The dominant production approach, exemplified by vLLM~\cite{kwon2023vllm}, binds one model per engine instance and handles multi-model deployments by running separate instances behind a request router, leaving model weights static in GPU memory and managing memory pressure solely through KV cache eviction or recomputation.
MuxServe~\cite{duan2024muxserve} relaxes this constraint by colocating multiple LLMs on the same GPU through spatial partitioning of streaming multiprocessors via NVIDIA MPS, combined with temporal interleaving of prefill and decode phases.
However, co-residency is hard-limited by GPU memory capacity: with typical model footprints of 20--30\,GB in FP16, at most two to three models can share an 80\,GB GPU simultaneously~\cite{xiang2025aegaeon}.
Prism~\cite{yu2025prism} introduces a two-level scheduler that places models across GPUs based on resource demands and dispatches requests to the assigned devices, but still requires loading and evicting models when the working set exceeds GPU memory.
ServerlessLLM~\cite{fu2024serverlessllm} introduces a loading-optimized checkpoint format and a multi-tier storage hierarchy that reduces cold-start latency by 10--200$\times$ compared to prior serverless frameworks, though models must still be fully loaded and unloaded for each switch.
Most recently, Aegaeon~\cite{xiang2025aegaeon} introduces token-level auto-scaling that dynamically loads and unloads models at fine granularity, reducing the GPU fleet needed to serve tens of models by 82\% in a production deployment at Alibaba Cloud.
Across all of these approaches, transferring model weights into and out of GPU memory remains the irreducible cost of multi-model serving, whether the source is disk, host DRAM, or a pre-staged CPU buffer.

A common assumption across the literature is that preemption carries a non-negligible cost, yet the actual magnitude of that cost, and its decomposition into constituent steps, has not been empirically isolated.
Kim et al.~\cite{kim2024dbms} rely on simulated overhead rather than measured transfer times; FastServe~\cite{wu2023fastserve} accounts for swap latency in its design but does not separate it from end-to-end throughput gains; FastSwitch~\cite{shen2024fastswitch} measures context-switch stalls but not the full unload--reload cycle.
Similarly, none of the offloading systems cited above characterize the continuous relationship between the fraction of layers offloaded and the resulting throughput degradation.
Our experiments address both gaps: Section~\ref{sec:offloading} sweeps the offloading ratio across models and hardware, and Section~\ref{sec:migrating} decomposes the full preempt--resume cycle into its constituent steps.
%
\section{Experimental Setup}
\label{sec:setup}
\subsection{Hardware}
The offloading experiment is conducted on two separate servers that share the same CPU subsystem, an AMD Threadripper PRO 5995WX (64 cores, 512\,GB DDR4) with a Samsung SSD 990 PRO 4TB, but differ in GPU hardware.
The first server is equipped with two NVIDIA RTX 5000 Ada Generation GPUs (32\,GB VRAM each, PCIe Gen\,4 x16), and the second with two NVIDIA RTX A6000 GPUs (48\,GB VRAM each, PCIe Gen\,4 x16).
The full offloading sweep is executed independently on each server to assess whether the qualitative behavior of layer offloading is consistent across GPU architectures while the CPU subsystem remains constant.
The preemption experiment uses a single GPU on each server.
Although both servers share the same PCIe Gen\,4 x16 interface, the two GPUs differ in memory controller behavior and driver-level transfer scheduling, leading to measurable differences in both model reload times and KV cache transfer latencies.
We therefore report preemption results separately for each platform to capture these hardware-dependent variations.

\subsection{Models and Software}

\Cref{tab:models} lists the models used across both experiments.
The offloading experiment serves three models through Ollama~v0.17.7 in their default quantized (Q4) formats: Llama~3~8B, Qwen3-32B, and Llama~2~70B.
With Ollama we control the number of transformer layers placed on the GPU, sweeping through target GPU fractions from 0\% (all CPU) to 100\% at 10\% increments, with additional fine-grained points at 92\%, 94\%, 96\%, and 98\% near full GPU residency.
Because percentage targets are rounded to integer layer counts, duplicate configurations are removed, yielding 13 to 15 unique placements per model depending on the total layer count.
The preemption experiment uses three models, Qwen2.5-3B, Qwen3-8B, and Qwen2.5-14B, loaded in FP16 directly via HuggingFace Transformers, which gives us programmatic access to the KV cache tensors needed for migration.
KV state is migrated between GPU and host memory using explicit device-to-device tensor copies over PCIe.
After each model unload, we force full GPU memory release and verify that allocated memory drops to driver-level residual before proceeding to the next step.

\begin{table}[]
\centering
\caption{Models used. VRAM columns show the quantized (Q4) size for the offloading experiment and the FP16 size for the preemption experiment. Layer counts correspond to Ollama layers (transformer blocks + output layer) for offloading and to transformer layers for preemption.}
\label{tab:models}
\begin{tabular}{lcccc}
\hline
\textbf{Model} & \textbf{Params} & \textbf{Layers} & \textbf{Q4} & \textbf{FP16} \\
\hline
\multicolumn{5}{l}{\textit{Offloading experiment}} \\
Llama 3 8B      & 8B  & 33 & 4.7\,GB  & ---      \\
Qwen3-32B        & 32B & 65 & 20\,GB   & ---      \\
Llama 2 70B      & 70B & 81 & 39\,GB   & ---      \\
\hline
\multicolumn{5}{l}{\textit{Preemption experiment}} \\
Qwen2.5-3B       & 3B  & 36 & ---      & 5.9\,GB  \\
Qwen3-8B         & 8B  & 36 & ---      & 15.6\,GB \\
Qwen2.5-14B      & 14B & 48 & ---      & 28.3\,GB \\
\hline
\end{tabular}
\end{table}

\subsection{Workloads}

The offloading experiment crosses 3~models $\times$ 13--15~unique GPU placements $\times$ 6~output lengths (50, 150, 300, 500, 1{,}000, and 5{,}000 tokens) $\times$ 3~repeats, totalling 774 inference calls per server.
The model is unloaded and reloaded before every call to ensure a clean layer placement.
The entire sweep is repeated on both servers to enable cross-hardware comparison.

The preemption experiment generates 7{,}000 tokens (Job~A) and interrupts it at nine checkpoints ($N \in \{100, 200, 300, 500, 1{,}000, 2{,}000, 3{,}000, 4{,}000, 5{,}000\}$ tokens), running a 500-token Job~B during each interruption.
Four model pairings are tested: Qwen2.5-3B$\to$Qwen3-8B, Qwen3-8B$\to$Qwen3-8B, Qwen3-8B$\to$Qwen2.5-14B, and Qwen2.5-14B$\to$Qwen3-8B, each with 2~repeats and compared against an uninterrupted baseline.
Job~A uses greedy decoding throughout.

\subsection{Metrics}

For partial offloading, we report decode throughput (tokens per second, tok/s) as a function of the fraction of layers placed on the GPU.
This metric captures the steady-state performance of autoregressive generation, where decoding dominates end-to-end latency for sufficiently long outputs.
By varying the number of offloaded layers, we quantify the trade-off between reduced GPU memory usage and the corresponding throughput degradation due to slower CPU execution and additional data movement.
We also report normalized throughput relative to the fully GPU-resident baseline, which allows direct comparison across models and hardware platforms.

For preemption, we instrument the execution pipeline by measuring the duration of each step in the preempt-resume cycle: KV cache transfer from GPU to CPU, model unload, replacement model load and execution, original model reload, and KV cache restoration from CPU to GPU.
We report the total preemption overhead, defined as the sum of these transfer and swap times, as well as its decomposition into model swap and KV transfer components.
To contextualize the absolute cost, we quantify the overhead as a percentage of the uninterrupted baseline completion time.
We further record the KV cache size in bytes at each checkpoint and measure effective PCIe bandwidth separately for each transfer direction (GPU-to-CPU and CPU-to-GPU).

%

\section{Layer-Wise CPU-GPU Offloading}
\label{sec:offloading}
The LLM inference can be performed on CPUs, GPUs, or hybrid configurations. In hybrid settings, the model parameters are partitioned across devices, with some layers stored in GPU memory and others in CPU memory. During inference, computation proceeds sequentially through the layers, and each layer is executed on the device where its parameters reside. When execution transitions between layers located on different devices, intermediate activations are transferred across the CPU-GPU interconnect to enable continuation of the forward pass.

\begin{figure*}[]
\centering
\includegraphics[width=0.9\linewidth]{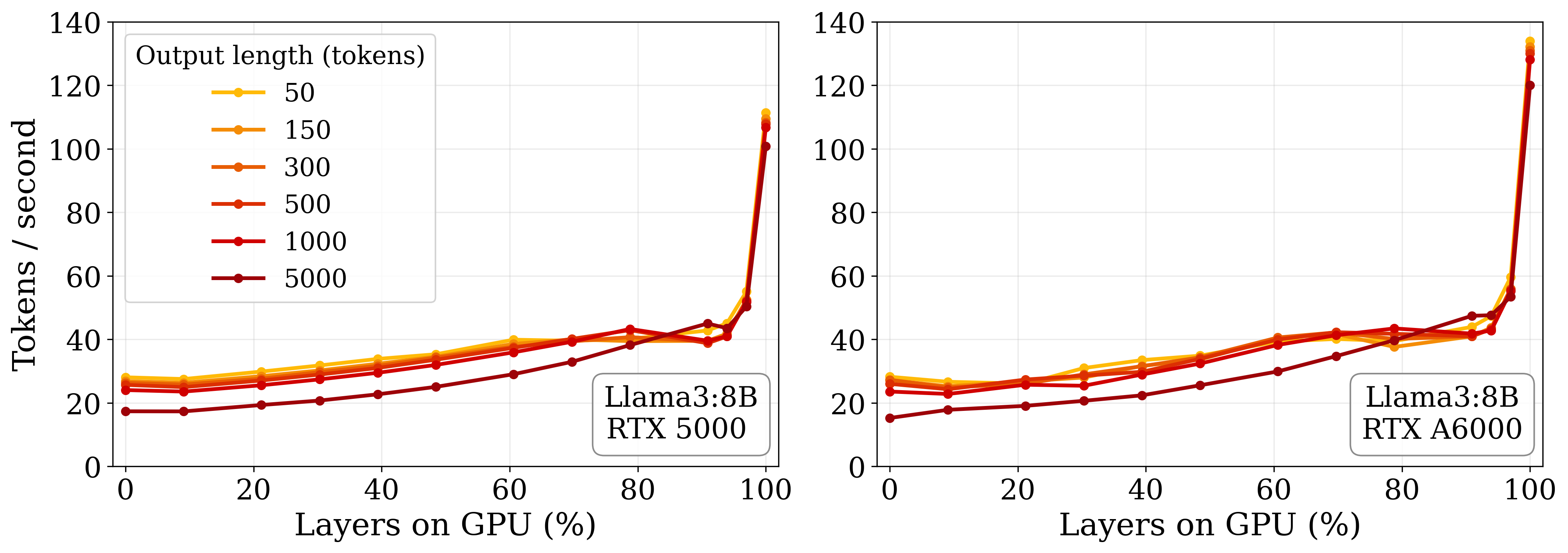}
\caption{Throughput by the Layer allocation of GPU \% for Llama3:8B model.}
\label{fig:Th_offload_llama8b}
\end{figure*}

\begin{figure*}[]
\centering
\includegraphics[width=0.9\linewidth]{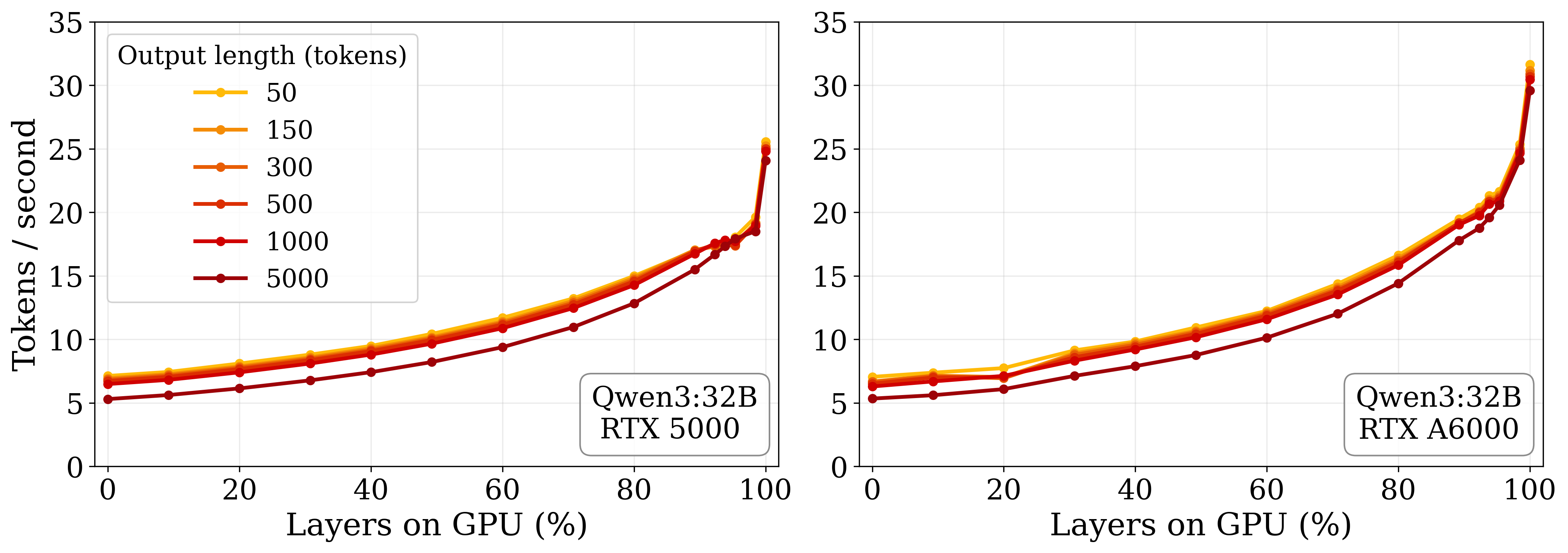}
\caption{Throughput by the Layer allocation of GPU \% for Qwen3:32B model.}
\label{fig:Th_offload_qwen32b}
\end{figure*}

\begin{figure*}[]
\centering
\includegraphics[width=0.9\linewidth]{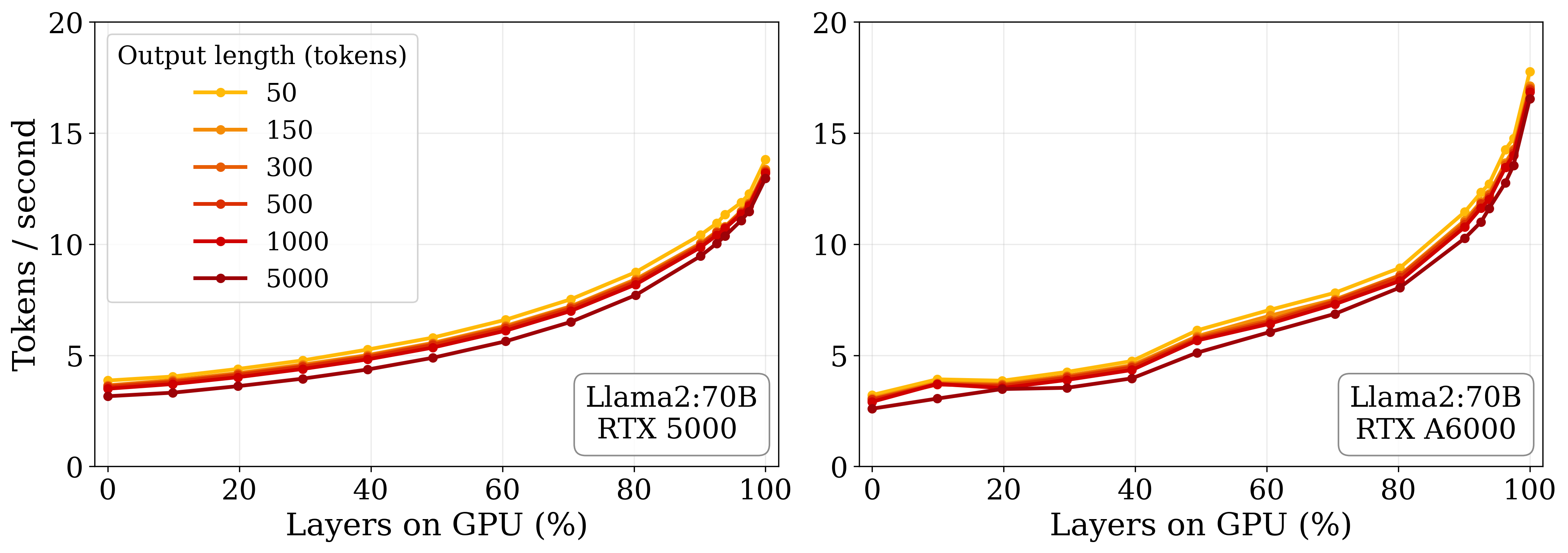}
\caption{Throughput by the Layer allocation of GPU \% for Llama2:70B model.}
\label{fig:Th_offload_llama70b}
\end{figure*}

Figure \ref{fig:Th_offload_llama8b}, \ref{fig:Th_offload_qwen32b}, \ref{fig:Th_offload_llama70b} report the throughput (tokens per second) depending on the fraction of layers that are executed on the GPU. The figures report different models of varying sizes under different prompt lengths on two different GPU hardware. 

A first observation is that, across all model sizes, throughput exhibits a consistent inverse relationship with prompt length, with longer prompts yielding proportionally lower throughput. This behavior arises from the growth of the KV cache, whose memory footprint scales linearly with the total sequence length. As the prompt length increases, both memory bandwidth pressure and cache access costs rise, thereby limiting effective token generation throughput.

Additionally, the scaling behavior varies with model size. Smaller models show a sharp increase in throughput as the fraction of layers placed on the GPU approaches 100\%, indicating a strong sensitivity to full GPU usage and a significant penalty from partial offloading. In contrast, larger models exhibit a more gradual, approximately linear improvement in throughput as GPU allocation increases, suggesting that incremental offloading yields smoother and more predictable performance gains. Although the GPUs differ in architectural characteristics and absolute performance, these qualitative trends are consistently observed across all evaluated devices.

\begin{figure*}[]
    \centering
    
    \subfigure[Llama3:8B]{\includegraphics[width=.31\textwidth]{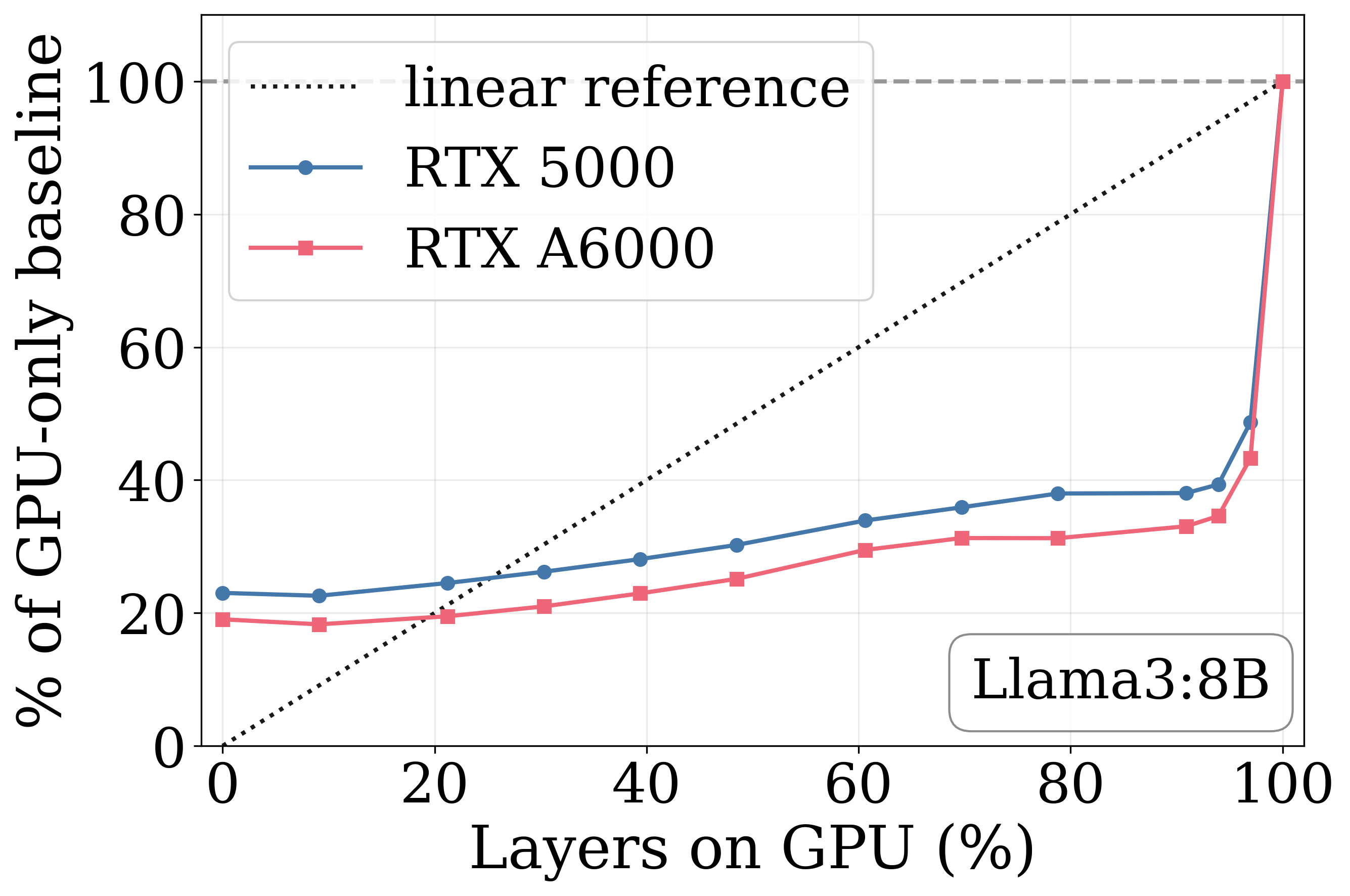}}\;
    \subfigure[Qwen3:32B]{\includegraphics[width=.31\textwidth]{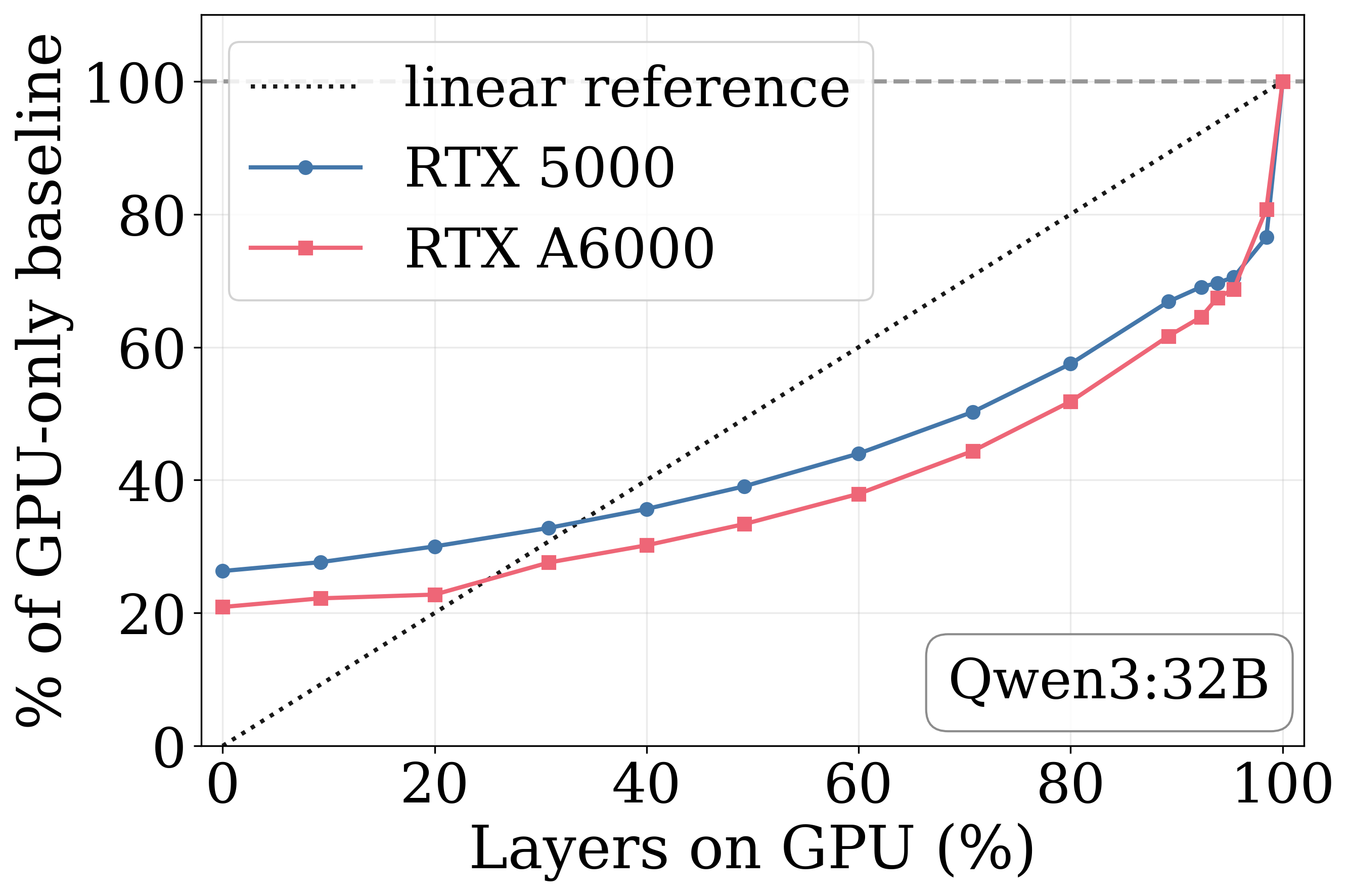}}\;
    \subfigure[Llama2:70B]{\includegraphics[width=.31\textwidth]{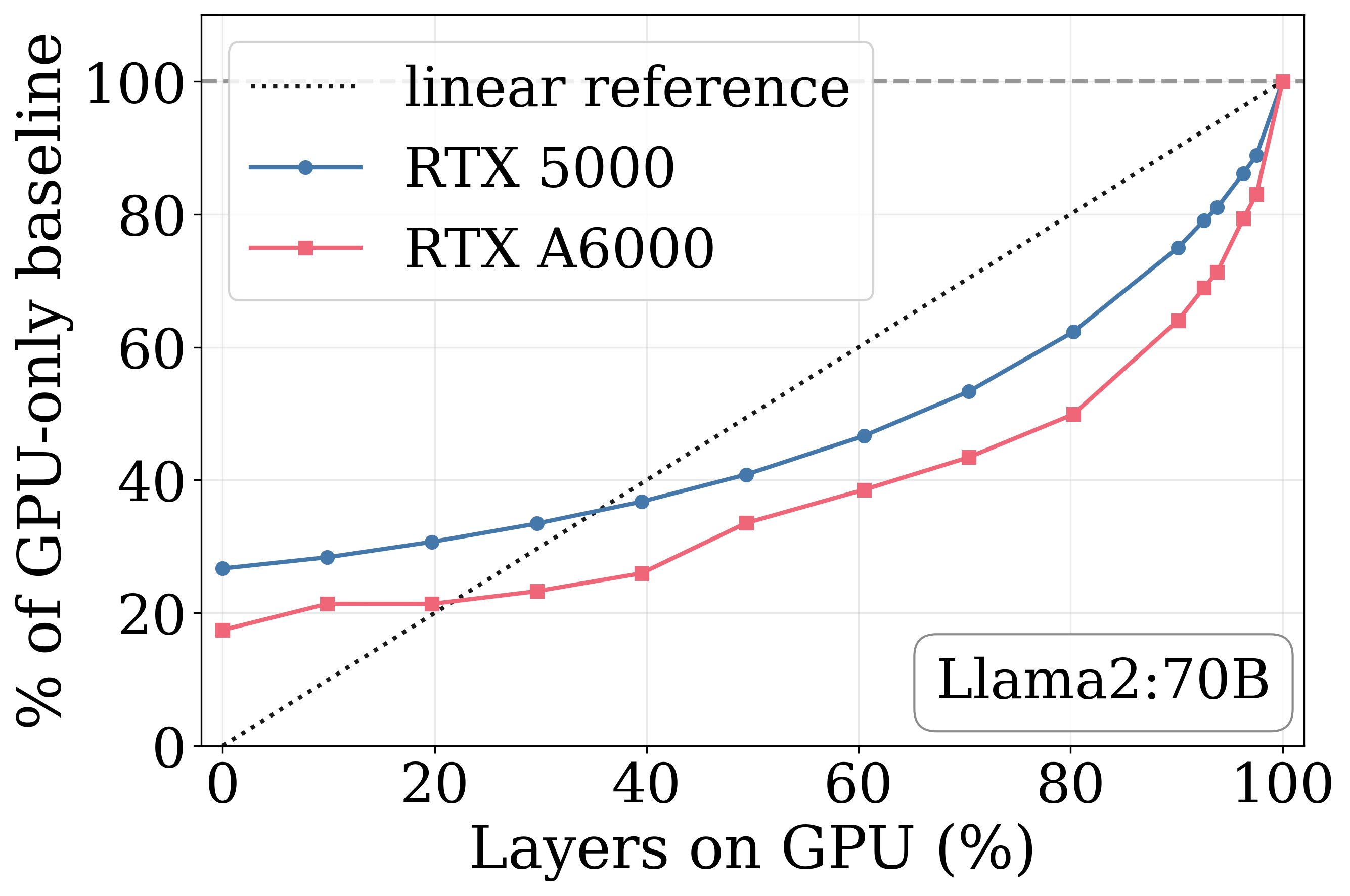}}\;

    \caption{Normalized throughput.}
    \label{fig:modsimnorm}
    \vspace{-0.35cm}
\end{figure*}

Figure \ref{fig:modsimnorm} presents inference throughput normalized to the 100\% GPU baseline, where all model parameters reside on the GPU and computation is fully GPU-bound. As observed previously, larger models exhibit behavior that more closely approaches linear scaling with respect to the fraction of layers placed on the GPU.

A second notable trend is that normalized throughput is consistently higher on the RTX 5000 compared to the RTX A6000 under partial offloading. This effect arises from the smaller performance gap between the CPU and the RTX 5000. Because the RTX 5000 has lower peak throughput, transferring a portion of the computation to the CPU incurs a smaller relative performance penalty. In contrast, the RTX A6000’s substantially higher compute capability amplifies the relative cost of CPU offloading, leading to greater degradation in normalized throughput.

In this scenario, the relative performance of the GPU and CPU must be jointly considered. The effectiveness of offloading depends not only on the absolute speed of the GPU, but also on its performance gap with respect to the CPU. Systems with high-end GPUs may experience disproportionately larger penalties from CPU offloading, whereas more balanced configurations, where CPU and GPU throughput are closer, can sustain higher normalized performance under hybrid execution. Consequently, scheduling policies should be hardware-aware, explicitly incorporating both GPU capability and CPU throughput to make optimal placement decisions.

%

\section{Dynamic KV Cache Migration}
\label{sec:migrating}
In this section, we benchmark the cost of full job preemption: pausing a running inference request, evicting its model and cached state from the GPU to service a higher-priority job, and then resuming the original request from where it left off.
Unlike partial offloading, which trades throughput for co-residency, preemption assumes exclusive GPU access for each job but pays a switching cost every time control changes hands.
We restrict our analysis to a single preemption per job; the aggregate cost under repeated preemption depends on the scheduling policy and is left to future work.
The central question is how large that one-time cost actually is, and which step of the preempt-resume cycle dominates the overhead.

We answer it with a controlled experiment on both hardware platforms described in \Cref{sec:setup}: the NVIDIA RTX 5000 Ada Generation (32\,GB VRAM) and the NVIDIA RTX A6000 (48\,GB VRAM), using a single GPU on each server.
A long-running \textit{Job~A} generates 7{,}000 tokens using greedy decoding in FP16.
At a designated checkpoint, after $N \in \{100, 200, 300, 500, 1000, 2000, 3000, 4000, 5000\}$ tokens, we preempt Job~A by copying its KV cache tensors to host memory over PCIe, then fully unloading its weights via explicit deletion and GPU cache flushing.
A second model, \textit{Job~B}, is loaded onto the free GPU and generates 500 tokens. 
Once Job~B completes, its model is unloaded in the same manner, Job~A's weights are reloaded from disk, the saved KV cache is restored from CPU to GPU, and generation resumes from the saved state to produce the remaining $7{,}000 - N$ tokens.
Comparing Job~A's total wall time against an uninterrupted baseline isolates the overhead introduced by preemption.

Four configurations cover different model-size pairings to isolate the effect of each model's footprint on preemption cost. 
The first run uses Qwen2.5-3B (5.9\,GB in FP16) as the preempted model (Job~A), providing a lightweight baseline. 
The second and third runs promote Qwen3-8B (15.6\,GB) to Job~A, with Job~B being Qwen3-8B and Qwen2.5-14B (28.3\,GB) respectively. 
The fourth run reverses the roles, preempting the larger Qwen2.5-14B with Qwen3-8B. 
Across these four pairings, the preempted model spans nearly a 5$\times$ range in memory footprint, from 5.9 to 28.3\,GB, letting us disentangle the effect of the preempted model's size from that of the preempting model.

The sequence is deliberately conservative: we fully unload each model rather than attempting to keep both resident, which would only be possible when their combined footprint fits within VRAM. 
The results, therefore, represent a worst-case preemption cost for memory-constrained hardware.

\begin{figure*}[]
\centering

    \subfigure[RTX 5000]{\includegraphics[width=\columnwidth]{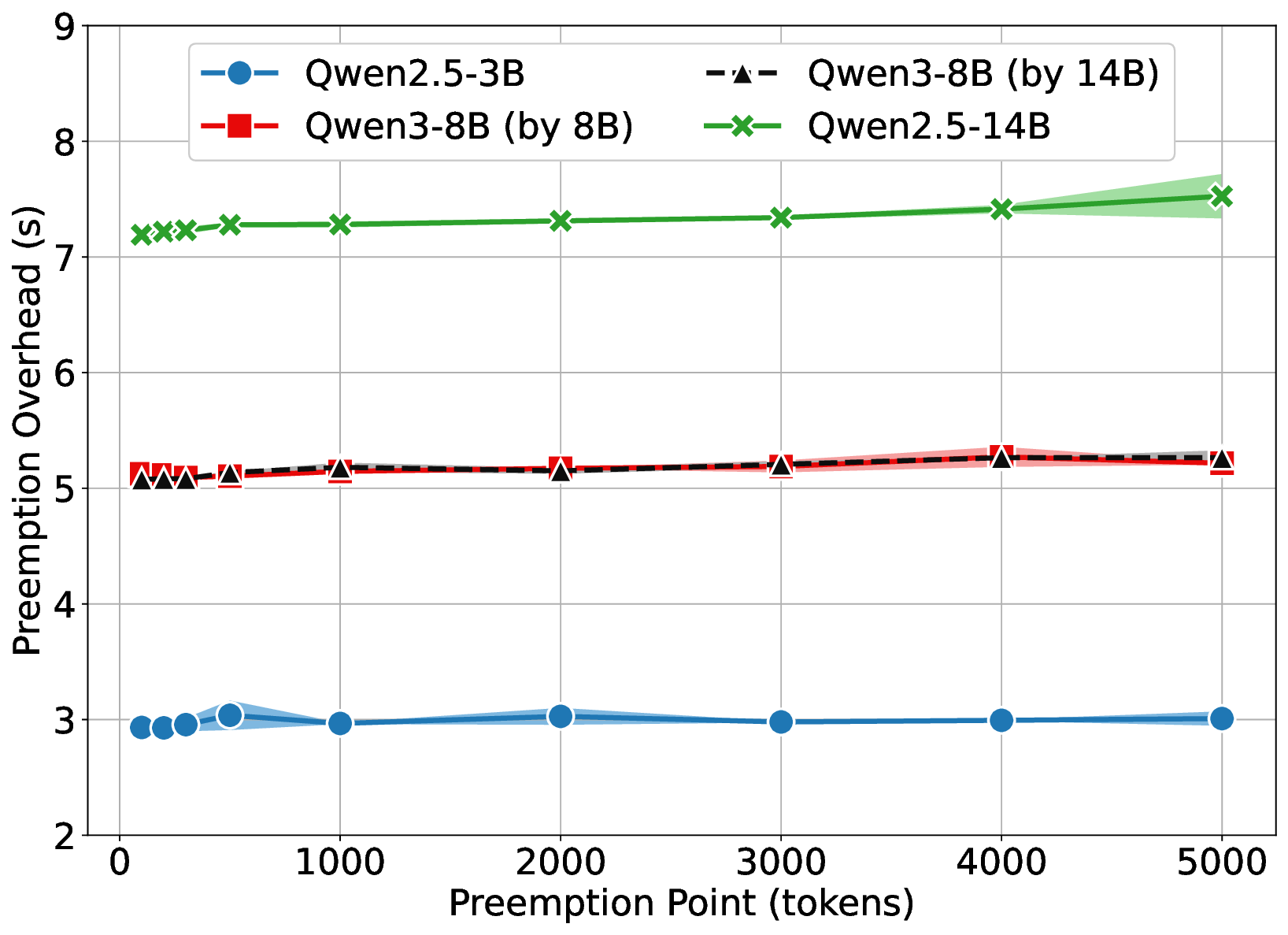}}\;
    \subfigure[RTX A6000]{\includegraphics[width=\columnwidth]{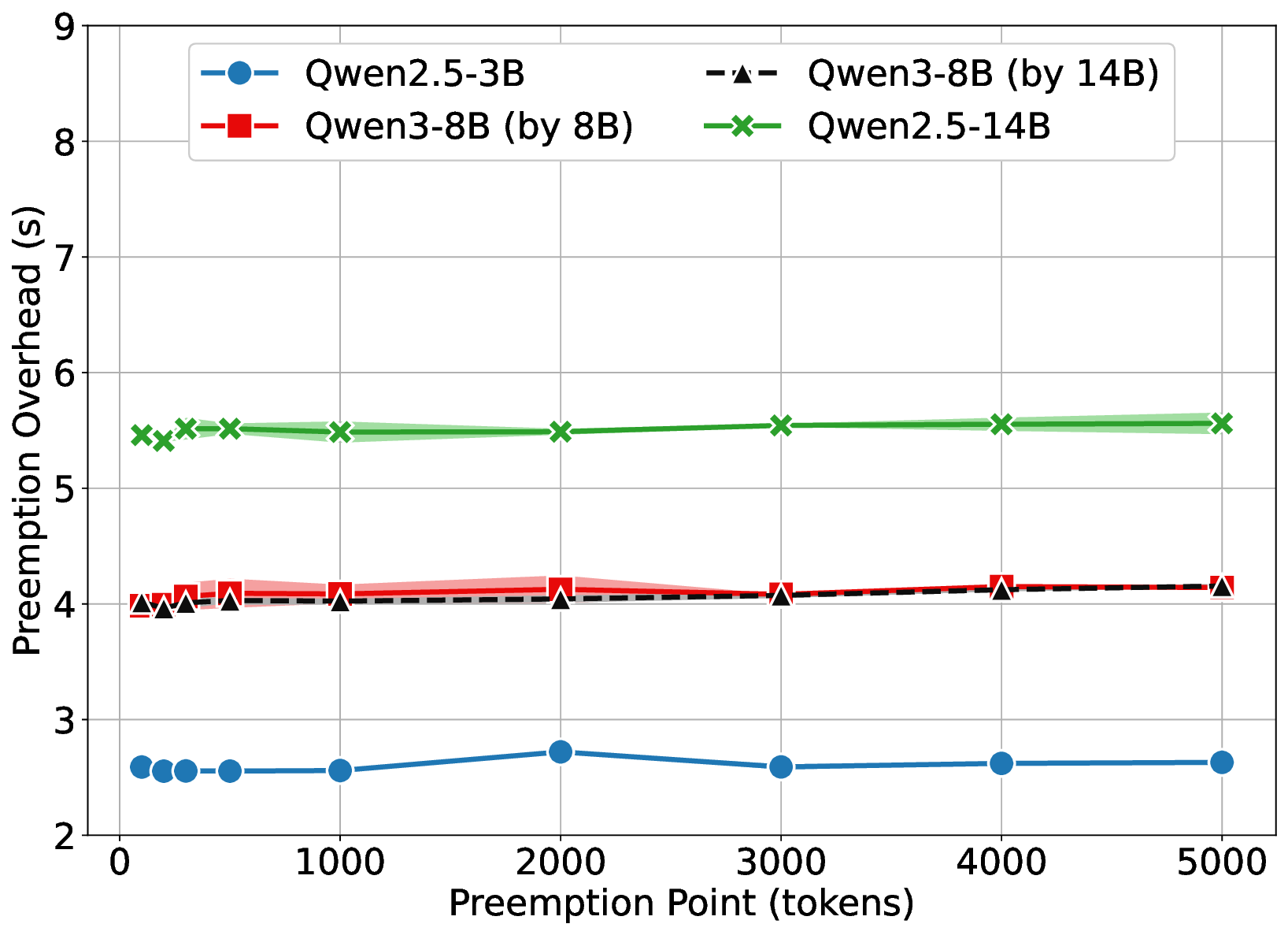}}\;

\caption{Preemption overhead vs.\ preemption point for four model configurations. 
Shaded bands indicate one standard deviation.}
\label{fig:overhead_vs_checkpoint}
\end{figure*}

\Cref{fig:overhead_vs_checkpoint} plots the total preemption overhead as a function of the preemption point (after how many tokens are generated) for all four configurations on both GPUs.
The central finding is immediate: every curve is effectively almost flat on both platforms, meaning that preemption overhead is constant regardless of how far the job has progressed at the time of interruption.
On the RTX 5000, Qwen2.5-3B incurs approximately 3\,s per preemption, both Qwen3-8B configurations settle around 5.1\,s, and Qwen2.5-14B reaches approximately 7.3\,s.
On the RTX A6000, the same qualitative pattern holds but with consistently lower absolute overhead: Qwen2.5-3B drops to approximately 2.6\,s, Qwen3-8B to around 4.1\,s, and Qwen2.5-14B to roughly 5.7\,s, reflecting faster disk-to-GPU transfer on that platform.
In neither case do these values shift appreciably, whether the job is interrupted after 100 or 5{,}000 tokens.
The shaded bands, which indicate one standard deviation across experimental repeats at each checkpoint, are narrow for the two smaller models and only slightly wider for Qwen2.5-14B, confirming that the result is not only flat but also robust across both hardware platforms.
A second observation follows from comparing the two Qwen3-8B curves: the solid line (preempted by Qwen3-8B) and the dashed line (preempted by Qwen2.5-14B) are nearly indistinguishable on both GPUs, indicating that the size of the preempting model (Job~B) has no measurable effect on the overhead experienced by the preempted job.
What determines the cost is the preempted model's own weight footprint, not the identity of the job that displaces it.

Since the size of the preempting model has no measurable effect on overhead, we consolidate the two Qwen3-8B runs into a single average and report results per preempted model only.
\Cref{tab:preemption_overhead} breaks the total overhead into its four constituent steps, averaged across all checkpoints, for both GPUs.
The breakdown reveals why overhead is constant: model reload dominates overwhelmingly.
On the RTX 5000, reload accounts for 2.71\,s, 4.70\,s, and 6.70\,s for Qwen2.5-3B, Qwen3-8B, and Qwen2.5-14B, respectively; on the RTX A6000, the corresponding values are 2.33\,s, 3.62\,s, and 5.12\,s, consistently lower due to faster disk-to-GPU transfer on that platform.
Adding the unload step, which is comparatively fast but still model-size-dependent, brings the model swap share to over 99\% of total overhead on the RTX 5000 and above 98.5\% on the RTX A6000 in every configuration.
Unloading is not instantaneous because it involves three sequential operations: Python's cyclic garbage collector must traverse the entire module tree to detect and break reference cycles among the thousands of parameter tensors and sub-modules that constitute the model, the PyTorch caching allocator must then return all freed memory blocks to the CUDA driver via individual deallocation calls, and finally a device synchronization must complete before the released memory becomes available to the next model.
Since both the number of objects to traverse and the number of cached blocks to free grow with model size, unload time scales from 0.26--0.28\,s for Qwen2.5-3B to 0.53--0.54\,s for Qwen2.5-14B across the two GPUs, with negligible difference between platforms as unloading is primarily a CPU-side operation.
The KV cache transfers in both directions combined never exceed 1\% of the total on the RTX 5000 and remain below 1.5\% on the RTX A6000, contributing tens of milliseconds at most.
This composition explains the flat curves in \Cref{fig:overhead_vs_checkpoint}: the dominant cost, reloading the model weights from disk into GPU memory, is a fixed function of the model's size and is entirely independent of how many tokens have been generated.

\begin{table*}[t]
\centering
\caption{Preemption overhead breakdown, averaged across all checkpoints and repeats.}
\label{tab:preemption_overhead}
\setlength{\tabcolsep}{4pt}
\begin{tabular}{l cc cc cc}
\toprule
& \multicolumn{2}{c}{\textbf{Qwen2.5-3B} (5.9\,GB)} 
& \multicolumn{2}{c}{\textbf{Qwen3-8B} (15.6\,GB)} 
& \multicolumn{2}{c}{\textbf{Qwen2.5-14B} (28.3\,GB)} \\
\cmidrule(lr){2-3} \cmidrule(lr){4-5} \cmidrule(lr){6-7}
& RTX 5000 & A6000 & RTX 5000 & A6000 & RTX 5000 & A6000 \\
\midrule
KV GPU $\to$ CPU (s)         & 0.0096 & 0.0125 & 0.0325 & 0.0387 & 0.0414 & 0.0439 \\
Unload Model A (s)           & 0.26   & 0.28   & 0.41   & 0.39   & 0.54   & 0.53   \\
Reload Model A (s)           & 2.71   & 2.33   & 4.70   & 3.62   & 6.70   & 5.12   \\
KV CPU $\to$ GPU (s)         & 0.0076 & 0.0075 & 0.0211 & 0.0205 & 0.0276 & 0.0275 \\
\midrule
Total overhead (s)           & 2.98   & 2.62   & 5.16   & 4.06   & 7.31   & 5.73   \\
Model swap (\%)              & 99.4   & 99.2   & 99.0   & 98.5   & 99.1   & 98.8   \\
KV transfer (\%)             & 0.6    & 0.8    & 1.0    & 1.5    & 0.9    & 1.2    \\
Overhead vs.\ baseline (\%)  & 2.04   & 1.72   & 2.11   & 1.86   & 1.75   & 1.61   \\
\bottomrule
\end{tabular}
\end{table*}

\begin{figure*}[ht]
\centering
\includegraphics[width=0.9\textwidth]{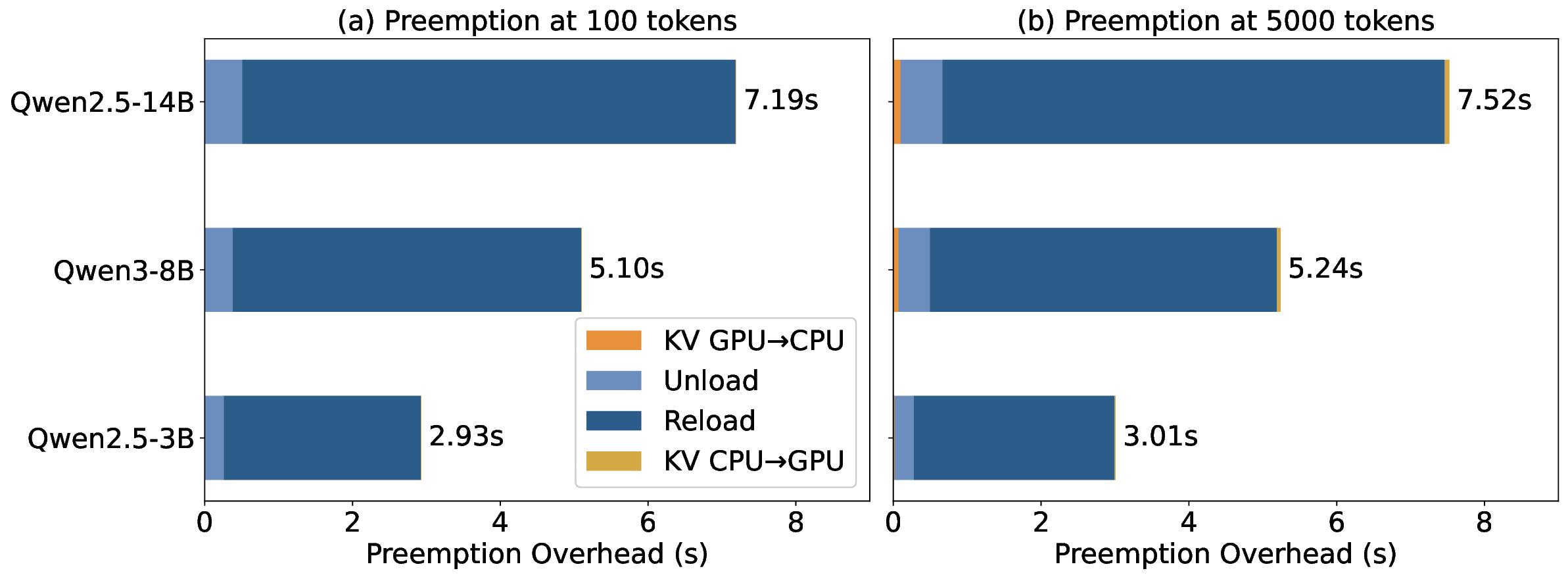}
\caption{Overhead composition at the earliest and latest preemption points for different models. The GPU used is RTX 5000.}
\label{fig:overhead_decomposition_RTX_5000}
\end{figure*}

\begin{figure*}[ht]
\centering
\includegraphics[width=0.9\textwidth]{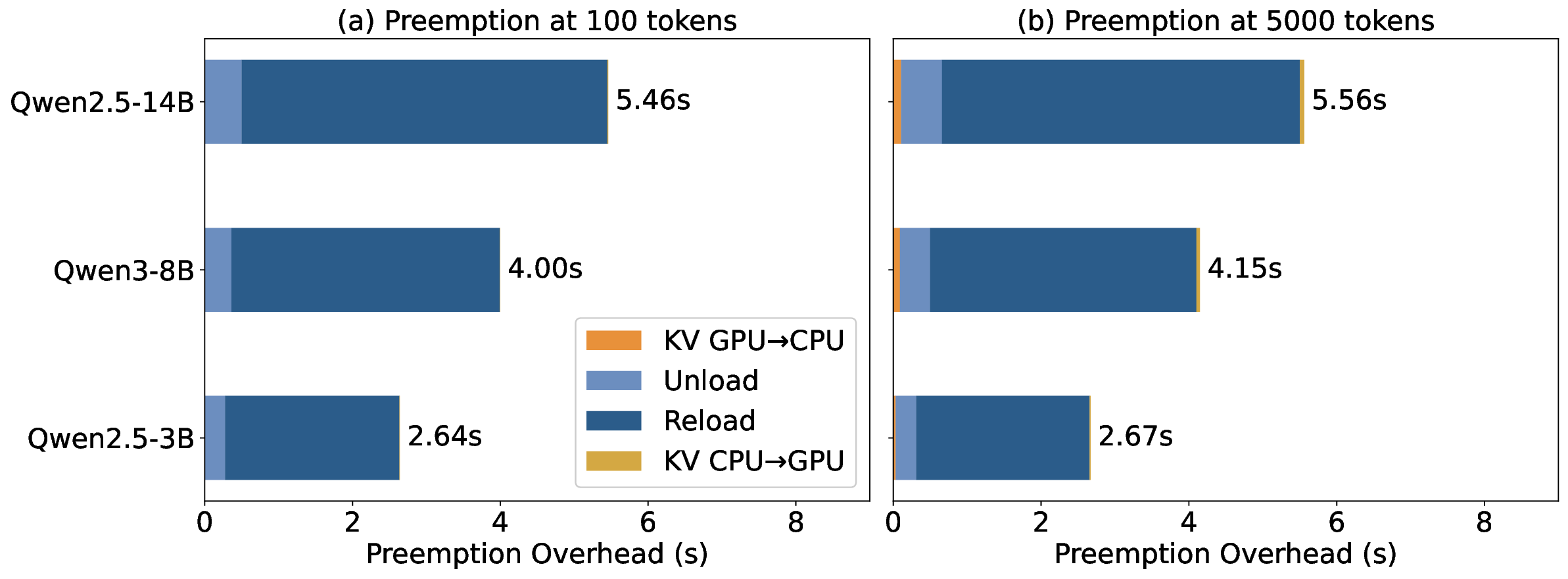}
\caption{Overhead composition at the earliest and latest preemption points for different models. The GPU used is RTX A6000.}
\label{fig:overhead_decomposition_RTX_A6000}
\end{figure*}

The last row of the table contextualizes the absolute cost in terms of end-to-end completion time.
On the RTX 5000, Qwen2.5-3B completes 7{,}000 tokens in 147.5\,s without interruption and in approximately 150\,s with a single preemption, an overhead of roughly 2\%.
Qwen3-8B goes from a 244\,s baseline to about 249\,s, again around 2\%.
The largest model, Qwen2.5-14B, rises from 417\,s to approximately 424\,s, an overhead of 1.7\%.
On the RTX A6000, the relative overhead is even smaller: 1.72\% for Qwen2.5-3B, 1.86\% for Qwen3-8B, and 1.61\% for Qwen2.5-14B, consistent with the lower absolute preemption cost on that platform.
In every case and on both GPUs, the preempted job finishes within seconds of the uninterrupted baseline, confirming that full preemption, even in this worst-case unload-reload scenario, is a lightweight operation whose cost is largely invisible at the job level.
\Cref{fig:overhead_decomposition_RTX_5000,fig:overhead_decomposition_RTX_A6000} visualize this composition at the two extremes of the checkpoint range on the RTX 5000 and RTX A6000, respectively.
At 100 tokens (left panels), the KV cache is just 6\,MB for Qwen2.5-3B, 24\,MB for Qwen3-8B, and 32\,MB for Qwen2.5-14B, and the orange and yellow slivers representing KV transfers are almost invisible on both platforms.
At 5{,}000 tokens (right panels), the KV cache has grown by roughly 30$\times$ to 178\,MB, 713\,MB, and 951\,MB respectively, yet the bars are virtually identical in length.
On the RTX 5000: 2.93\,s versus 3.01\,s for Qwen2.5-3B, 5.10\,s versus 5.24\,s for Qwen3-8B, and 7.19\,s versus 7.52\,s for Qwen2.5-14B.
On the RTX A6000 the same pattern holds at lower absolute values: 2.64\,s versus 2.67\,s for Qwen2.5-3B, 4.00\,s versus 4.15\,s for Qwen3-8B, and 5.46\,s versus 5.56\,s for Qwen2.5-14B.
The blue segment, model reload, visually overwhelms every other component in all panels across both GPUs, making it clear that even a 30$\times$ increase in the data that must be shuttled across PCIe produces no meaningful change in total overhead.

Although KV cache transfer contributes less than 1\% of total preemption overhead on the RTX 5000 and below 1.5\% on the RTX A6000, it is the only component whose cost grows with the length of the interrupted job.
Characterizing this growth is important for two reasons: it reveals how efficiently the PCIe link is utilized during migration, and it establishes the regime in which KV transfer would begin to matter, for instance, on future hardware with faster model loading or for models serving very long contexts where caches reach tens of gigabytes.
\Cref{tab:kv_transfer} reports the KV cache size and per-direction transfer time at each checkpoint for both GPUs.

\begin{table*}[t]
\centering
\caption{KV cache size and transfer time across checkpoints.}
\label{tab:kv_transfer}
\begin{tabular}{ll c cc cc}
\toprule
\textbf{Model} & \textbf{Tokens} & \textbf{KV} 
& \multicolumn{2}{c}{\textbf{RTX 5000}} 
& \multicolumn{2}{c}{\textbf{RTX A6000}} \\
\cmidrule(lr){4-5} \cmidrule(lr){6-7}
& & (MB) & GPU$\to$CPU & CPU$\to$GPU & GPU$\to$CPU & CPU$\to$GPU \\
& &      & (ms) & (ms) & (ms) & (ms) \\
\midrule
             & 100  &   6.0 &  4.4 &  2.8 &  5.4 &  3.0 \\
             & 500  &  20.1 &  5.7 &  4.8 &  5.8 &  5.0 \\
Qwen2.5-3B  & 1000 &  37.7 &  8.2 &  7.9 &  7.3 &  8.0 \\
             & 2000 &  72.8 & 11.8 &  8.6 & 11.4 &  8.5 \\
             & 3000 & 108.0 & 12.9 & 10.0 & 18.0 &  9.9 \\
             & 5000 & 178.3 & 18.0 & 14.4 & 31.3 & 14.1 \\
\midrule
             & 100  &  24.0 &  6.1 &  6.1 &  5.6 &  6.0 \\
             & 500  &  80.3 & 12.0 &  9.0 & 11.2 &  8.9 \\
Qwen3-8B    & 1000 & 150.6 & 18.7 & 13.5 & 28.5 & 13.7 \\
             & 2000 & 291.2 & 33.0 & 22.3 & 50.4 & 22.8 \\
             & 3000 & 431.9 & 45.7 & 30.4 & 65.7 & 30.5 \\
             & 5000 & 713.1 & 67.8 & 45.9 & 90.3 & 45.4 \\
\midrule
             & 100  &  32.1 &  8.7 &  7.9 &  7.0 &  9.5 \\
             & 500  & 107.1 & 16.0 & 12.9 & 14.9 & 12.7 \\
Qwen2.5-14B & 1000 & 200.8 & 29.7 & 18.6 & 30.7 & 18.6 \\
             & 2000 & 388.3 & 45.8 & 30.1 & 54.4 & 29.5 \\
             & 3000 & 575.8 & 62.1 & 41.1 & 70.5 & 39.7 \\
             & 5000 & 950.8 & 86.7 & 62.5 & 104.9 & 59.8 \\
\bottomrule
\end{tabular}
\end{table*}

The cache grows linearly with the number of generated tokens, at a rate that scales with model dimension: 0.035\,MB/token for Qwen2.5-3B, 0.14\,MB/token for Qwen3-8B, and 0.19\,MB/token for Qwen2.5-14B.
These rates are intrinsic to the model architecture and are therefore identical across both GPUs; only the transfer times differ.
At the largest checkpoint (5{,}000 tokens), the cache reaches 178\,MB, 713\,MB, and 951\,MB respectively, yet even the largest transfer completes in under 90\,ms in the GPU-to-CPU direction and under 63\,ms in the reverse direction on the RTX 5000.
On the RTX A6000, GPU-to-CPU times are moderately higher, reaching up to 105\,ms for Qwen2.5-14B at 5{,}000 tokens, while CPU-to-GPU times remain comparable at under 60\,ms.
The GPU-to-CPU path is consistently slower than CPU-to-GPU on both platforms because CUDA blocks device-to-host transfers until the copy has fully completed, whereas host-to-device transfers may return before the DMA operation finishes~\cite{cuda2025progguide}.
Effective bandwidth rises with transfer size as the fixed per-transfer overhead is amortized: at 5{,}000 tokens the GPU-to-CPU path reaches 10--12\,GB/s and the CPU-to-GPU path reaches 13--16\,GB/s on the RTX 5000, with the RTX A6000 achieving similar CPU-to-GPU rates but somewhat lower GPU-to-CPU throughput (9--10\,GB/s), both well below the 31.5\,GB/s theoretical peak of a PCIe Gen\,4 x16 link due to the overhead of per-layer tensor copies through the PyTorch runtime.
These numbers confirm that, even without optimized bulk-transfer implementations, KV cache migration is a fast operation whose cost is dwarfed by the time needed to reload model weights from disk on both platforms.
Taken together, these results yield a simple rule of thumb for scheduling: the cost of preempting an LLM inference job can be modeled as a fixed penalty that depends only on the weight footprint of the preempted model, not on how much progress the job has made.
On our PCIe Gen\,4 hardware, this penalty ranges from roughly 3\,s for a 3B-parameter model to 7\,s for a 14B-parameter model on the RTX 5000, and from approximately 2.6\,s to 5.7\,s on the RTX A6000, adding no more than 2\% to job completion time for long-running jobs such as the 7{,}000-token generation used in this experiment.
KV cache migration, despite growing linearly with sequence length, remains negligible in comparison and would only become significant if model reload times were reduced by an order of magnitude.
These findings simplify the design of preemption-aware schedulers, which can treat the swap cost as a known constant per model and hardware pair rather than a variable that depends on the work done.

%

\section{Discussion and Feature Set Identification}
\label{sec:discussion}

The experimental results highlight that scheduling LLM inference in heterogeneous CPU-GPU environments is fundamentally governed by a set of interacting factors that extend beyond simple load balancing.
In this section, we distill these observations into a set of key features that future schedulers must explicitly consider when supporting partial offloading and preemption.

A first critical feature is the non-linear relationship between the fraction of layers placed on the GPU and the resulting throughput.
Across all evaluated models, decode throughput does not scale linearly with the percentage of layers assigned to the GPU, deviating significantly from the linear reference.
This effect is particularly pronounced for smaller models such as Llama3:8B or Qwen3:32B, where even moderate CPU offloading leads to substantial degradation in normalized performance, whereas larger models exhibit a flatter degradation curve and greater tolerance to heterogeneous execution.
This implies that schedulers must incorporate model-specific sensitivity to offloading, rather than relying on proportional or uniform placement strategies.

A second feature is the strong dependence on model architecture, beyond raw parameter count.
The results show clear differences between models of similar scale, such as Qwen and Llama, indicating that architectural choices influence how computation and memory are distributed across layers and therefore affect both throughput and offloading efficiency.
Consequently, scheduling decisions must be aware of model-specific execution profiles, rather than treating all models as interchangeable workloads.

A third feature is the interaction between offloading and sequence length.
Throughput measurements across varying output lengths indicate that longer sequences amplify the impact of offloading, as decode-phase inefficiencies accumulate over time.
While prefill costs are amortized, decode throughput becomes the dominant factor, and any slowdown introduced by CPU execution is repeatedly incurred.
This suggests that schedulers must account for request characteristics, particularly expected output length, when deciding placement and resource allocation.

A fourth feature concerns the cost structure of preemption.
The decomposition of preemption overhead reveals that the dominant component is not KV cache transfer, but rather model state reload, which accounts for over 99\% of total overhead on the RTX 5000 and above 98.5\% on the RTX A6000.
This cost remains relatively stable across different preemption points, indicating that preemption overhead is largely insensitive to when it occurs during generation.
However, the absolute overhead varies significantly across both models and hardware: from roughly 3\,s for a 3B-parameter model to 7\,s for a 14B-parameter model on the RTX 5000, and from 2.6\,s to 5.7\,s on the RTX A6000.
This highlights that preemption is not a lightweight operation and must be used selectively, with awareness of both model-dependent and hardware-dependent costs.

A fifth feature is the role of KV cache size and data movement.
Although KV transfer contributes less than 1.5\% of total preemption overhead on both platforms, it is the only component that grows with the length of the interrupted job, and it becomes increasingly relevant for longer sequences where the cache scales linearly with token count.
The measured PCIe bandwidth indicates that interconnect limitations can become a bottleneck, particularly in scenarios involving frequent preemption or aggressive offloading.
Therefore, schedulers must consider both compute placement and data movement costs, especially under high concurrency or long-context workloads.

Finally, the results indicate that hardware characteristics significantly influence all of the above factors.
Differences between GPUs (e.g., RTX 5000 vs. RTX A6000) affect not only the baseline throughput and the relative penalty of offloading, but also the absolute cost of preemption, with the RTX A6000 exhibiting lower reload times despite its larger VRAM capacity.
This reinforces that scheduling policies must be hardware-aware, treating the swap cost as a known constant per model and hardware pair, and adapting decisions to the specific compute, memory, and I/O capabilities of the underlying devices.

Taken together, these observations suggest that effective scheduling for multi-model LLM inference requires a holistic approach that integrates model-aware, workload-aware, and hardware-aware features.
Future schedulers should explicitly model (i) non-linear offloading sensitivity, (ii) model architecture and scale, (iii) sequence length and decode dynamics, (iv) preemption overhead decomposition, and (v) data movement constraints.
Incorporating these features is essential to enable efficient preemption and hybrid CPU-GPU execution in realistic, heterogeneous deployment scenarios.

%

\section{Conclusions}
\label{sec:conclusions}

In this paper, we presented a fine-grained empirical study of CPU-GPU layer offloading and job-level preemption for LLM inference, deriving a set of design features for schedulers targeting multi-model LLM deployments.
Our measurements expose two regimes that current single-model schedulers do not capture.
First, decode throughput degrades non-linearly with partial offloading, and the shape of this degradation depends on both the model and the GPU, so a single offloading curve cannot drive placement decisions.
Second, the cost of a full preempt-resume cycle is effectively almost constant regardless of how far the interrupted job has progressed, since model reload from disk dominates the overhead while KV cache migration remains a minor contribution even at long sequence lengths.
Together, these results allow preemption-aware schedulers to treat the swap cost as a known, fixed penalty per model and hardware pair, and the features distilled in Section~\ref{sec:discussion} translate these regimes into concrete inputs that a multi-model scheduler can act on.

As future work, the offloading and preemption characterization should be extended to continuous-batching scenarios with concurrent requests, since all experiments in this study use single-request execution.
The preemption analysis should also be generalized beyond a single interruption per job, as the aggregate cost under repeated preemptions will depend on the scheduling policy and may exhibit different trade-offs.
A workload-level study is needed to estimate how often a typical inference job would need to be preempted under realistic arrival patterns, and to compare the resulting performance against fully non-preemptive policies, in order to determine whether preemption yields a net benefit or a net loss on average.

\section*{Acknowledgment}
This work was supported by the European Union - Next Generation EU under the Italian National Recovery and Resilience Plan (NRRP), Mission 4, Component 2, Investment 1.3, CUP B53C22004050001, partnership on “Telecommunications of the Future” (PE00000001 - program “RESTART”)

\bibliographystyle{IEEEtran}
\bibliography{references}

\end{document}